\documentclass[12pt]{spieman}  
\usepackage{amsmath,amsfonts,amssymb}
\usepackage{graphicx}
\usepackage{setspace}
\usepackage{tocloft}
\usepackage{lineno}
\usepackage{tabularx}
\usepackage{booktabs}

\title{Dual-View Optical Flow for 4D Micro-Expression Recognition - A Multi-Stream Fusion Attention Approach}

\author[a]{Luu Tu Nguyen}
\author[a]{Thi Bich Phuong Man}
\author[a]{Vu Tram Anh Khuong}
\author[a]{Thanh Ha Le}
\author[a,*]{Thi Duyen Ngo}
\affil[a]{VNU University of Engineering and Technology, Faculty of Information Technology, Hanoi, Vietnam}

\cftpagenumbersoff{figure}
\cftpagenumbersoff{table} 
\begin{document} 
\maketitle

\begin{abstract}
Micro-expression recognition is vital for affective computing but remains challenging due to the extremely brief, low-intensity facial motions involved and the high-dimensional nature of 4D mesh data. To address these challenges, we introduce a dual-view optical flow approach that simplifies mesh processing by capturing each micro-expression sequence from two synchronized viewpoints and computing optical flow to represent motion. Our pipeline begins with view separation and sequence-wise face cropping to ensure spatial consistency, followed by automatic apex-frame detection based on peak motion intensity in both views. We decompose each sequence into onset-apex and apex-offset phases, extracting horizontal, vertical, and magnitude flow channels for each phase. These are fed into our Triple-Stream MicroAttNet, which employs a fusion attention module to adaptively weight modality-specific features and a squeeze-and-excitation block to enhance magnitude representations. Training uses focal loss to mitigate class imbalance and the Adam optimizer with early stopping. Evaluated on the multi-label 4DME dataset, comprising 24 subjects and five emotion categories, in the 4DMR IJCAI Workshop Challenge 2025, our method achieves a macro-UF1 score of 0.536, outperforming the official baseline by over 50\% and securing first place. Ablation studies confirm that both the fusion attention and SE components each contribute up to 3.6 points of UF1 gain. These results demonstrate that dual-view, phase-aware optical flow combined with multi-stream fusion yields a robust and interpretable solution for 4D micro-expression recognition.
\end{abstract}


{\noindent \footnotesize\textbf{*} Thi Duyen Ngo,  \linkable{duyennt@vnu.edu.vn} }

\begin{spacing}{2}   

\section{Introduction}

\noindent Micro-Expressions (MEs) are involuntary, brief, and difficult-to-detect facial movements that often occur when people try to hide their true emotions. Unlike regular expressions, which can be faked or controlled, micro-expressions genuinely reflect inner emotions and are therefore of increasing interest in areas such as criminal investigation, negotiation, and behavioral analysis \cite{Bhushan2015}\cite{yan2013fast}\cite{polikovsky2010detection}.

Micro-expression recognition (MER) is a challenging problem in emotion analysis due to its low intensity and extremely short duration. Traditional methods based on visual data often use 2D features such as LBP-TOP\cite{zhao2007dynamic} or HOG\cite{1467360} on videos, which have achieved some promising results but still face difficulties in capturing subtle spatio-temporal movements, especially under varying lighting conditions or head poses. Recent deep learning models, including CNNs and RNNsn\cite{review1}\cite{10.1007/978-3-031-27066-6_2}, have shown potential in exploiting spatial and temporal information, but they mainly rely on 2D data or landmark sequences, which are insufficient to fully represent facial dynamics\cite{article55, xia2019spatiotemporalrecurrentconvolutionalnetworks, CNNCapsNet}. In recent years, multi-stream or multi-view architectures have shown potential in combining different types of features, such as mesh geometry, landmark displacement vectors, or optical flow (OF) computed from 2D videos. However, these methods still face limitations when dealing with complex micro-movements under varying lighting conditions, viewpoints, and poses, and they have not yet fully exploited the dynamic information that 4D data can provide. To overcome these limitations, 4D (3D + time) data has been explored as a new direction for MER. 3D mesh sequences allow detailed descriptions of facial geometric changes over time, thereby capturing micro-movements that are difficult to observe in 2D space. However, the effective utilization of this data presents significant challenges.

First and foremost is the computational trade-off between dimensionality and efficiency. Directly processing raw 4D data (dynamic 3D meshes or point clouds) requires complex geometric deep learning architectures (e.g., PointNet++ \cite{qi2017pointnetdeephierarchicalfeature} or 3D CNNs \cite{tran2015learningspatiotemporalfeatures3d}), which impose a prohibitive computational burden and memory footprint. This complexity hinders the deployment of such systems in real-world scenarios where computational resources may be limited. Therefore, there is a critical need for a method that can leverage the rich depth information of 4D data while maintaining the computational efficiency of 2D CNNs. Our approach addresses this by projecting the 4D spatiotemporal data into synchronized dual-view optical flow maps, effectively reducing the dimensionality while preserving the essential geometric motion cues.

Secondly, micro-expressions are characterized by their extremely low intensity and short duration. In traditional RGB-based or raw mesh-based approaches, the subtle pixel intensity changes or vertex shifts caused by micro-expressions are often overshadowed by redundant information such as facial identity or static textures. By converting the sequence into optical flow streams (horizontal, vertical, and magnitude), we explicitly decouple motion from appearance. This ensures that the network focuses solely on the dynamic evolution of facial muscles rather than static facial characteristics.

Finally, existing multi-stream architectures typically treat different motion components equally, often employing naive concatenation for feature fusion. However, different micro-expressions manifest differently across motion directions (e.g., eyebrow raising is predominantly vertical, while a smirk may involve significant horizontal stretching). A static fusion strategy fails to capture this nuance. To overcome this, we propose an adaptive multi-stream fusion mechanism. By integrating a Fusion Attention module, our model dynamically learns to recalibrate the importance of Horizontal, Vertical, and Magnitude streams, ensuring that the most discriminative motion features are emphasized for each specific expression instance.

Overall, this paper proposes a 4D micro-expression recognition method based on Dual-view optical flow, realized through a triple-stream deep learning architecture with a fusion attention mechanism. The core of the method is to exploit three types of robust motion features (the u, v, and magnitude components of optical flow) computed from 3D mesh sequences, and feed them into parallel streams to model different aspects of micro-expressions. These streams are then combined by an attention module, enabling the system to emphasize important signals and suppress noise. In addition, we investigate data augmentation techniques specifically designed for 4D sequences, such as apex-off augmentation, to improve the generalizability and robustness of the model.

Our main contributions include:
\begin{itemize}
    \item Proposing MicroAttNet, a three-branch architecture with a unified attention mechanism, specifically designed for MER on 4D data.
    \item Efficient extraction and analysis of 4D optical flow features (u, v, and magnitude) from 3D mesh sequences.
    \item Introducing a suitable data augmentation strategy for 4D micro-expressions.
    \item Evaluating the proposed method on the 4DMR Challenge dataset\cite{9796028} – the first 4D dataset to record spontaneous micro-expressions with high-quality mesh sequences – and achieving outstanding results.
\end{itemize}

Finally, our method is validated in the international 4D Micro-expression recognition for Mind Reading (4DMR 2025) challenge, where it ranked first place on the official leaderboard, confirming the effectiveness and potential for practical applications of the proposed system.

\section{The 1st Challenge for 4D Micro-Expression Recognition for Mind Reading}
The goal of this challenge was to recognize micro-expressions using 4D data, focusing on benchmarking algorithms for detecting subtle, involuntary facial movements in spontaneous high-stakes scenarios. Held as part of the 1st Challenge and Workshop for 4D Micro-Expression Recognition for Mind Reading (4DMR 2025) in conjunction with IJCAI 2025, the event aimed to advance robust techniques leveraging 4D analysis (3D mesh + temporal dynamics) over traditional 2D/3D methods, addressing real-world challenges like illumination variations, head poses, occlusions, and environmental noise. The challenge was hosted on Kaggle (https://www.kaggle.com/competitions/4-dmr-ijcai-workshop-challenge-2025), with top-performing teams ranked by performance metrics. In this challenge, participants are tasked with recognizing five categories of micro-expressions: Negative, Positive, Repression, Surprise, Others. The task is framed as a multi-label classification problem, where a single expression instance may exhibit features from multiple emotion categories. This better reflects the complexity of real human emotions and provides a more nuanced benchmark for recognition systems. By the end of the challenge, there were a total of 17 participating teams and 234 submissions.
 
\section{Proposed method}
\begin{figure*}[ht]
    \centering
    \includegraphics[width=0.95\textwidth]{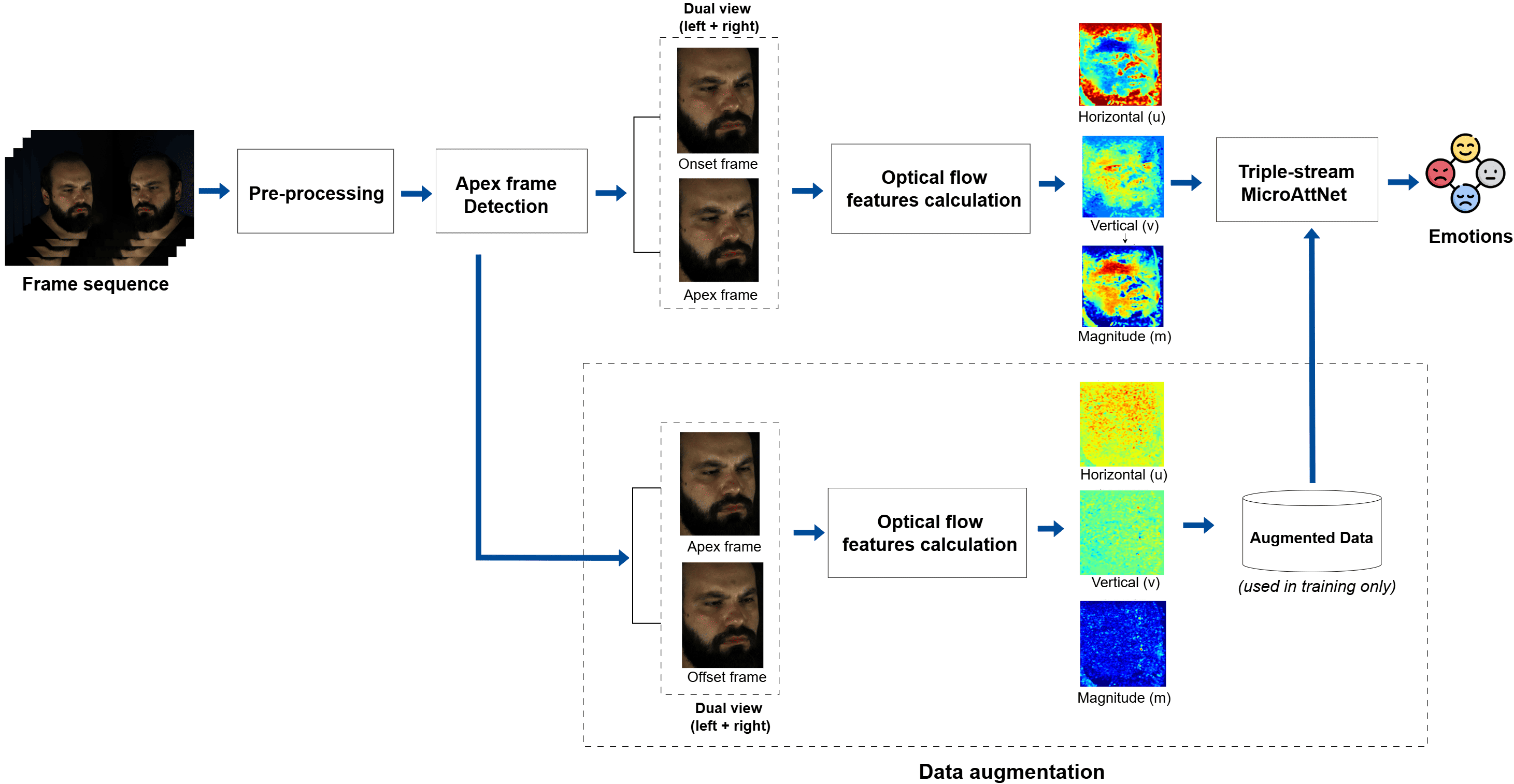}
    \caption{The workflow of our proposed method}
    \label{fig:pipline}
\end{figure*}

The proposed MER method employs a dual-view, triple-stream attention-based network for micro-expression recognition. Specifically, optical flow is computed independently for the left and right facial views, each represented using three motion channels - horizontal ($u$), vertical ($v$), and magnitude ($m$). These 3-channel optical flow features are then processed through a dedicated attention mechanism to capture phase-aware and view-consistent motion patterns. The overall workflow of the proposed approach is illustrated in Fig.~\ref{fig:pipline}.

\subsection{Pre-processing}
Although the 4DME dataset provides 3D (3D + temporal) data, we adopt a 2D-based approach for this challenge. This choice is motivated by the fact that the texture images (.jpg) in 4DME already contain high-resolution facial expression details, which are sufficient for micro-expression analysis. Moreover, processing in 2D enables the application of a wide range of well-established computer vision techniques, while being computationally more efficient. Importantly, since micro-expressions involve subtle muscle movements without significant changes in head pose, a 2D representation still preserves the critical information necessary for recognition.

Each input image contains two facial views: the left and right sides of the subject's face. To utilize both views effectively, we split each input image into two separate images, one for the left face and one for the right face. Following this split, we apply face cropping to isolate the facial region in each image, focusing on the areas where micro-expressions are most likely to occur. To maintain spatial consistency across frames within a sequence, we define a fixed cropping rule: for each sequence (sample), we first identify the bounding box of the face in every frame, and then select the bounding box with the largest area as the standard cropping region. This bounding box is subsequently applied to all frames within the sequence. This approach ensures uniform face positioning and size throughout the sample, which is essential for stable model training and evaluation. Given that micro-expressions involve minimal head movement and primarily affect facial muscles, our cropping strategy enhances alignment and reduces potential variability caused by small shifts or frame-to-frame differences. 

\begin{figure}[ht]
    \centering
    \includegraphics[width=0.88\linewidth]{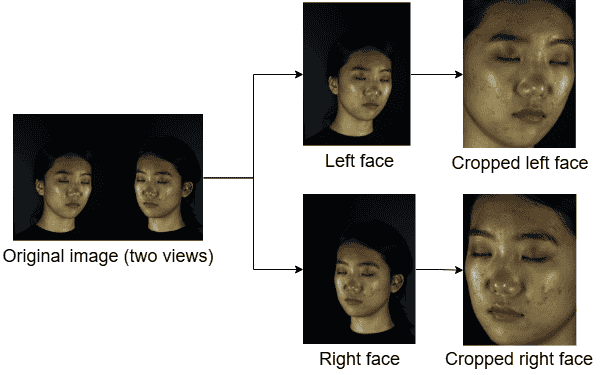}
    \caption{The preprocessing pipeline for 4DME dataset}
    \label{fig:preprocess}
\end{figure}

The complete preprocessing pipeline, including view separation, face detection, bounding box selection, and sequence-wise cropping, is illustrated in Fig.~\ref{fig:preprocess}.

\subsection{Apex frame detection}
The apex frame, indicating the peak of facial muscle activity, is essential for capturing the temporal dynamics of micro-expressions. Since the 4DME dataset in this challenge lacks ground-truth apex annotations, an automatic detection method based on motion intensity is employed.

Each sequence is assumed to start from a neutral state, with the first frame designated as the onset frame. Dense optical flow is calculated between the onset and subsequent frames using the Farneback algorithm \cite{farneback}. For each frame $f$, the flow magnitude at pixel $(x,y)$ is computed as follows:

\begin{equation}
\label{eq:magnitude}
M(x, y) = \sqrt{u(x, y)^2 + v(x, y)^2}
\end{equation}

where \(u(x, y)\) and \(v(x, y)\) represent the horizontal and vertical components of the flow at pixel \((x, y)\), respectively.

The total motion intensity is defined as the sum of flow magnitudes across all pixels:

\begin{equation}
\label{eq:motion}
I_f = \sum_{x, y} \sqrt{u(x, y)^2 + v(x, y)^2}
\end{equation}

The apex frame is then identified as the one with the highest intensity:

\begin{equation}
\label{eq:result}
    \text{apex\_frame} = \arg \max_{f \in \text{Frames}} I_f
\end{equation}

The apex frame estimation is performed independently for both the left and right facial views, ensuring phase-aligned motion representation across the two perspectives. This motion-based approach enables reliable and unsupervised apex localization, enhancing the discriminative power of the optical flow features used for recognition.

\subsection{Optical flow features calculation}
After detecting apex frame, we extract optical flow representations to characterize the subtle facial motion across two distinct temporal segments: onset–apex and apex–offset. This phase-aware decomposition is grounded in the temporal nature of micro-expressions, which exhibit a bell-shaped motion profile, facial muscle intensity typically increases from onset to apex and then decreases towards offset. Modeling each phase separately enables the network to capture phase-specific motion patterns and directional asymmetries that are often overlooked when treating the sequence as a whole.

For each phase, dense optical flow is computed between the corresponding frame pairs using the Farneback method~\cite{farneback}. From the resulting flow fields, we extract three complementary motion components: Horizontal displacement ($u$), Vertical displacement ($v$) and Motion magnitude ($m$). We adopt the Farneback method because it provides dense optical flow estimates at a relatively low computational cost compared to newer but more computationally demanding methods (e.g., RAFT, TV-L1), while still capturing subtle local motion patterns that are important for ME analysis. Its polynomial expansion formulation is particularly effective in modeling small displacements between consecutive frames, which helps detect the subtle and transient motions characteristic of micro-expressions. These components reflect both the direction and intensity of facial motion. Rather than fusing them into a single input tensor, we design a triple-stream architecture where each stream independently processes one motion modality. This architectural choice encourages the learning of complementary spatial and motion-sensitive features, thereby improving the model’s ability to discriminate between subtle expression classes.

For each phase, dense optical flow is computed between the corresponding frame pairs using Farneback method~\cite{farneback}. From the resulting flow fields, we extract three complementary motion components: Horizontal displacement ($u$), Vertical displacement ($v$) and Motion magnitude ($m$). These components reflect both the direction and intensity of facial motion. Rather than fusing them into a single input tensor, we design a triple-stream architecture where each stream independently processes one motion modality. This architectural choice encourages the learning of complementary spatial and motion-sensitive features, thereby improving the model’s ability to discriminate between subtle expression classes.

\subsection{Phase-Aware Temporal Augmentation method}
Building upon our prior research on micro-expression dynamics \cite{phase}, we integrate a Phase-Aware Temporal Augmentation strategy to enhance model generalization. While this approach has previously demonstrated the ability to improve motion understanding in 2D contexts, this study marks its first successful application to 4D micro-expression data, validating its effectiveness on depth-aware optical flow features.

To further enhance temporal robustness, samples from both the \textit{onset-apex} (rising intensity) and \textit{apex-offset} (falling intensity) phases are jointly utilized during training. By explicitly incorporating the decaying phase of the expression, we not only increase the diversity of motion patterns observed by the network but also promote the learning of features that are invariant to the temporal position of informative cues. Consequently, the model is better equipped to handle variations in expression duration, subject identity, and phase alignment, ultimately leading to more effective classification performance.
\begin{figure*}[ht]
    \centering
    \includegraphics[width=0.95\linewidth]{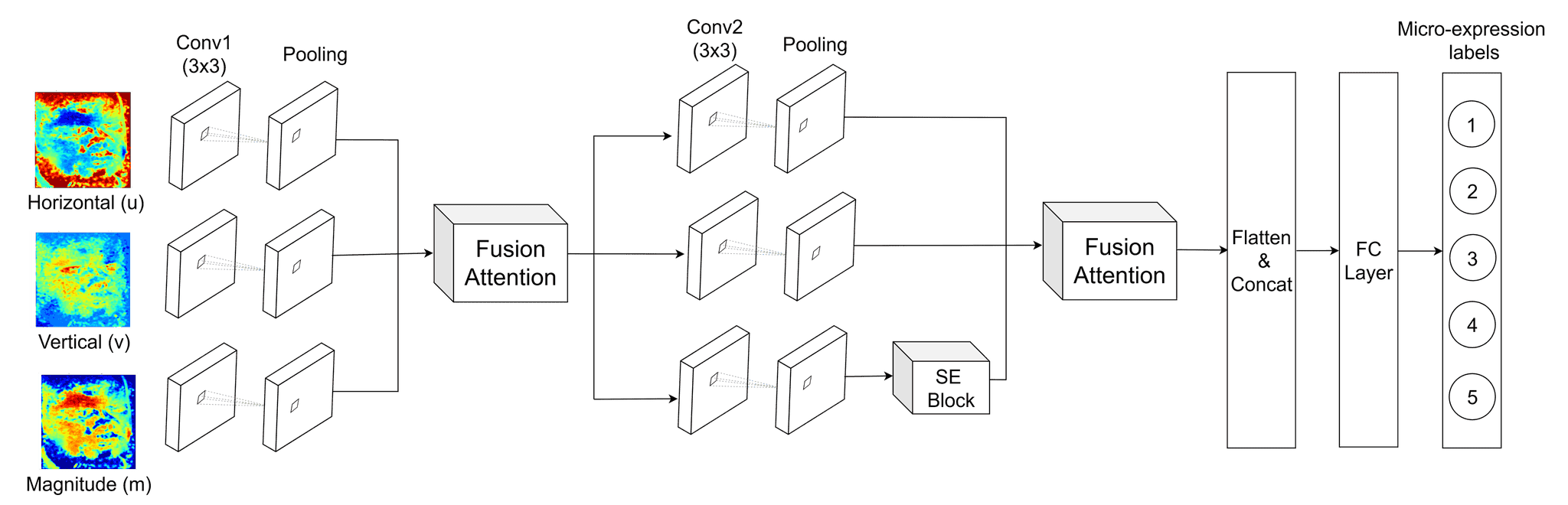}
    \caption{The architecture of Triple-stream MicroAttNet}
    \label{fig:network}
\end{figure*}
\subsection{Triple-stream MicroAttNet}

This paper introduces a lightweight triple-stream convolutional architecture tailored for micro-expression recognition from optical flow representations (see Fig.~\ref{fig:network}).
Let the input tensor be denoted as $X \in R^{B\times 3 \times H \times W}$. The input consists of three channels corresponding to the horizontal ($u$), vertical ($v$), and motion magnitude ($m$) components of optical flow, stacked into a 3-channel tensor: 
\[
X = \lbrace {X_h, X_v, X_m} \rbrace
\]
where: $X_h, X_v, X_m \in R^{B\times 1 \times H \times W}$ represent the horizontal optical flow, vertical optical flow, and optical flow magnitude, respectively.

Each stream, corresponding to the $u$, $v$ and $m$ motion components, is processed independently through two convolutional blocks.  Let $\mathbf{F}_k^{(l)}$ denote the feature map of stream $k \in \{h, v, m\}$ at layer $l$. The operation for a standard convolutional block is defined as:

$$\mathbf{F}_k^{(l)} = \text{Pool}\left(\text{Dropout}\left(\sigma\left(\text{BN}\left(\mathbf{W}_k^{(l)} * \mathbf{F}_k^{(l-1)} + \mathbf{b}_k^{(l)}\right)\right)\right)\right)$$
where: $*$ denotes the convolution operation.
$\text{BN}(\cdot)$ denotes batch normalization.
$\sigma(\cdot)$ denotes the ReLU activation function.
$\text{Pool}(\cdot)$ denotes max pooling.

\begin{figure*}[ht]
    \centering
    \includegraphics[width=0.75\linewidth]{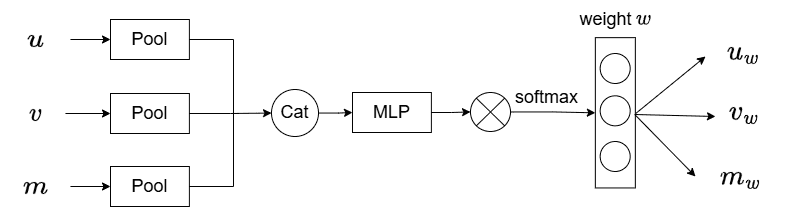}
    \caption{The Fusion Attention used in MicroAttNet. It adaptively weights the input streams ($u$, $v$, $m$) by generating soft attention weights, producing weighted outputs ($u_w$, $v_w$, $m_w$) that emphasize informative motion directions.}
    \label{fig:att}
\end{figure*}

To effectively integrate information across the three streams, we introduce a Fusion Attention module followed after each convolutional block, as illustrated in Fig.~\ref{fig:att}. Specifically, given input feature maps $\mathbf{H}, \mathbf{V}, \mathbf{M}$ from the three streams. \textbf{\textit{Global context aggregation}} extracts global descriptors using global average pooling:
$$g_k = \frac{1}{H \times W} \sum_{i=1}^{H} \sum_{j=1}^{W} \mathbf{F}_k(i,j), \quad k \in \{h, v, m\}$$

Followed by \textbf{\textit{attention weight generation}} module. The descriptors are concatenated ($\mathbf{g} = [g_h, g_v, g_m]$) and passed through a shared multi-Layer perceptron to learn adaptive weights:
$$\mathbf{z} = \mathbf{W}_2 \cdot \sigma(\mathbf{W}_1 \cdot \mathbf{g})$$
$$\boldsymbol{\alpha} = \text{Softmax}(\mathbf{z})$$
where: $\boldsymbol{\alpha} = [\alpha_h, \alpha_v, \alpha_m]$ represents the attention weights for the horizontal, vertical, and magnitude streams, respectively.
This attention mechanism enables the network to dynamically prioritize the most informative motion direction at each stage of representation learning.
Finally, the feature maps are re-weighted channel-wise:
$$\tilde{\mathbf{F}}_k = \alpha_k \cdot \mathbf{F}_k, \quad k \in \{h, v, m\}$$

To further enhance the feature discriminability of the magnitude stream, a Squeeze-and-Excitation (SE) block~\cite{seblock} is incorporated after the second convolutional layer in the $m$ branch.
For the magnitude feature map $\mathbf{F}_m$:$$\mathbf{s} = \text{GAP}(\mathbf{F}_m)$$$$\mathbf{e} = \sigma_{sigmoid}\left(\mathbf{W}_{se2} \cdot \sigma\left(\mathbf{W}_{se1} \cdot \mathbf{s}\right)\right)$$$$\mathbf{F}_m^{SE} = \mathbf{F}_m \otimes \mathbf{e}$$where $\otimes$ denotes channel-wise multiplication.
 
The SE mechanism explicitly models inter-channel dependencies by applying global pooling followed by channel-wise gating through a bottleneck MLP. This recalibration strategy helps amplify informative channels while suppressing less relevant ones, thereby improving the representation of motion intensity encoded in the magnitude stream.

The outputs refined feature maps from all three streams are flattened and concatenated to form a unified representation vector:
$$\mathbf{V}_{final} = \text{Concat}\left( \text{Flatten}(\tilde{\mathbf{F}}_h^{(2)}), \text{Flatten}(\tilde{\mathbf{F}}_v^{(2)}), \text{Flatten}(\tilde{\mathbf{F}}_m^{(2)}) \right)$$

The final classification is performed via a fully connected layer with Dropout and Batch Normalization:$$\hat{\mathbf{y}} = \mathbf{W}_{out} \cdot \text{Dropout}\left(\sigma\left(\text{BN}\left(\mathbf{W}_{fc} \cdot \mathbf{V}_{final}\right)\right)\right)$$
The architecture employs dropout and batch normalization after each convolutional stage to improve generalization and training stability.

To address the class imbalance inherent in micro-expression datasets, the model is optimized using the Focal Loss \cite{lin2017focal}:
$$L(\mathbf{y}, \hat{\mathbf{p}}) = - \sum_{i=1}^{C} y_i (1 - \hat{p}_i)^\gamma \log(\hat{p}_i)$$where $y_i$ is the ground truth, $\hat{p}_i$ is the predicted probability, and $\gamma$ is the focusing parameter.

\section{Experiments and Results}
\label{s:experiment}
\subsection{Dataset}
This paper uses the 4DME dataset, also adopted in 4DMR 2025 \cite{9796028}, which contains 4D facial expression sequences capturing 3D spatial changes over time. Each sequence includes 3D meshes (.obj), material files (.mtl), and facial texture images (.jpg), focusing on segments with micro-expressions - subtle, fast, and involuntary facial movements.

The dataset has 24 subjects from diverse cultural backgrounds, with expressions labeled into five categories: Negative, Positive, Repression, Surprise, and Others, forming a multi-label classification task. In this study, only the .jpg images are used for feature extraction and model training, as they provide key texture and expression details for analyzing micro-expression dynamics.

\subsection{Evaluation Metrics}
The performance of models is evaluated using the F1-score, which balances precision and recall. It is calculated as follows:

\begin{equation}
F1 = 2 \cdot \frac{\text{precision} \cdot \text{recall}}{\text{precision} + \text{recall}}
\end{equation}

For this paper and 4DMR challenge, we use the \text{macro F1-score}, computed as the average F1-score across all emotion classes:

\begin{equation}
F1_{macro} = \frac{1}{N} \sum_{i=1}^{N} F1_i
\end{equation}

where $N$ is the number of classes and $F1_i$ is the F1-score for the $i^{th}$ class.

This metric ensures that all emotion classes are evaluated fairly, regardless of class imbalance. The macro F1-score is a combined measure of precision and recall, making it well-suited for micro-expressions, which are typically brief, subtle, and difficult to detect.

\subsection{Implementation Details}
All experiments are conducted using PyTorch. Inputs are resized to 224×224 and normalized. The models are trained using the Adam optimizer with a learning rate of $10^{-4}$ for 50 epochs, with early stopping based on validation loss. The batch size is set to 32. To address the class imbalance commonly observed in micro-expression datasets, focal loss~\cite{lin2017focal} is employed as the primary classification objective. Focal loss extends the standard cross-entropy loss by introducing a modulating factor that down-weights easy examples, thereby focusing the learning process on hard and underrepresented samples. 

\subsection{Results}
\label{s:result}
Table~\ref{tab:res} summarizes the leaderboard results of the 4DMR 2025. Our proposed method, submitted under the team name \textit{Red-Green-Blue}, achieved the highest UF1 score of 0.536, outperforming both the official baseline (UF1 = 0.355) and other competitive entries. This result demonstrates the effectiveness of our dual-view, phase-aware design for micro-expression recognition.

\begin{table}[ht]
\caption{The top 5 leaderboard of the 4DMR IJCAI Workshop Challenge 2025}
\label{tab:res}
\begin{center}
\begin{tabular}{@{}llr@{}}
\toprule
\textbf{Rank} & \textbf{Team} & \textbf{Score} \\
\midrule
1 & \textbf{Red-Green-Blue (ours)} & \textbf{0.536} \\
2 & WS & 0.518 \\
3 & Infinite Messtropy & 0.476 \\
4 & Mohammad Javad Shamani & 0.473 \\
5 & \textit{Benchmark} & \textit{0.355} \\
\midrule
1 & \textbf{MicroAttNet (ours late submission)} & \textbf{0.554} \\
\bottomrule
\end{tabular}
\end{center}
\end{table}

To arrive at this final configuration, we conducted a series of internal experiments to evaluate different architectural variants. As reported in Table~\ref{tab:ablation}, we performed ablation studies by removing key components from the proposed model, including the Fusion Attention module and the SE block applied to the magnitude stream. The results show that each component contributes positively to performance: removing Fusion Attention reduced UF1 to 0.524, while removing the SE block reduced it to 0.518. When both components were removed, performance dropped further to 0.510. The full configuration with both components enabled achieved the highest UF1 of 0.554, and was selected for submission to the challenge.

\begin{table*}[h]
\caption{Performance comparison of different model configurations}
\label{tab:ablation}
\begin{center}
\begin{tabular}{lccc}
\toprule
\textbf{Configuration} & \textbf{Fusion Attention} & \textbf{SE Block} & \textbf{UF1 Score} \\
\midrule
\textbf{MicroAttNet (ours)} & X & X & \textbf{0.554} \\
Without Fusion Attention &   & X & 0.524\\
Without SE Block & X &  & 0.518 \\
Without both &  &  & 0.510\\
\bottomrule
\end{tabular}
\end{center}
\end{table*}

\begin{figure}[ht]
    \centering
    \includegraphics[width=0.9\linewidth]{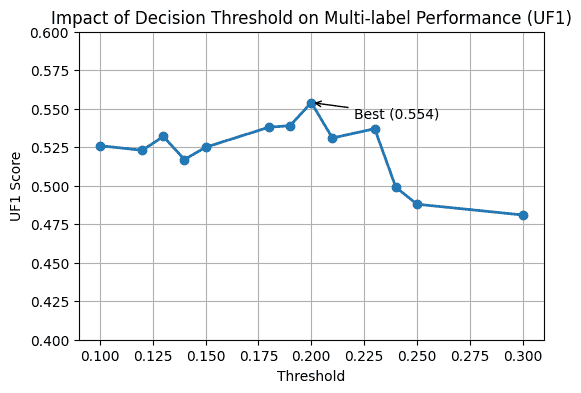}
    \caption{Impact of Decision Threshold on Multi-label Performance (UF1)}
    \label{fig:thresh}
\end{figure}

In addition to architectural choices, we also examined the effect of varying the decision threshold in the multi-label classification setting. As illustrated in Fig.~\ref{fig:thresh}, thresholds ranging from 0.10 to 0.30 were tested. The best UF1 score (0.554) was observed at a threshold of 0.20, although performance remained relatively stable in the range of 0.18 - 0.22. Based on these results, a threshold of 0.20 was chosen for final inference. These tuning efforts played a critical role in maximizing recognition accuracy and securing the top position on the challenge leaderboard.

\section{Practical experiences, strategic insights and findings of participated tasks}
Participation in the 4D Micro-Expression Recognition Challenge (4DMR 2025) provided important strategic insights. Although the task involves processing 4D data (3D + temporal), our approach - based on converting the data into 2D dual-view optical-flow motion features - demonstrated superior performance. Specifically, the Triple-Stream MicroAttNet architecture achieved a macro-UF1 score of 0.536 and secured first place, significantly outperforming other methods, including those that directly process 3D data. This confirms that extracting phase-robust 2D motion features (onset–apex, apex–offset) and integrating Fusion Attention is an efficient strategy in terms of both accuracy and computational cost for handling 4D data.

However, the achieved efficiency (UF1 $\approx$ 0.5x) also reflects the inherent complexity of the micro-expression recognition problem, which involves extremely short and low-intensity facial movements. To further improve performance, future developments are necessary. First, there is an urgent need to design more specialized deep models to fully capture the complex spatio-temporal correlations in MER. Second, the simultaneous use and processing of multiple types of data (multi-modality), including 3D mesh data in addition to 2D texture images, can provide more comprehensive and complementary information, thereby significantly improving the accuracy of micro-expression recognition.

\section{Conclusion}
\label{s:conclusion}
This paper has introduced a dual-view approach for micro-expression recognition from 4D facial data. By independently estimating apex frames for both left and right facial views, and projecting the 3D motion into a compact 2D representation using three-stream optical flow features (horizontal, vertical, and magnitude), the proposed method effectively captures subtle temporal dynamics. Extensive experiments on the 4DMR Challenge dataset demonstrate that our model achieves promising results, ranking first on the official leaderboard. These findings highlight the potential of combining multi-view facial cues with phase-specific motion modeling for advancing micro-expression recognition in high-dimensional settings.


\appendix    


\subsection*{Disclosures}
The authors declare that there are no financial interests, commercial affiliations, or other potential conflicts of interest that could have influenced the objectivity of this research or the writing of this paper.

\subsection* {Code, Data, and Materials Availability} 
Data sharing is not applicable to this article, as no new data were created or analyzed.



\bibliography{report}   
\bibliographystyle{spiejour}   



\listoffigures
\listoftables

\end{spacing}
\end{document}